\documentclass[11pt]{article}
\usepackage{aaai}
\usepackage{times}
\usepackage{latexsym}
\usepackage{amsmath}
\usepackage{multirow}
\usepackage{url}
\usepackage{graphics}

\usepackage{enumerate}

\title{Using Nuances of Emotion to Identify Personality}

 \author{Saif M. Mohammad and Svetlana Kiritchenko\\
  	National Research Council Canada\\
 	 Ottawa, Ontario, Canada K1A 0R6\\
   {\tt \{saif.mohammad,svetlana.kiritchenko\}@nrc-cnrc.gc.ca} }

\date{}

\begin{document}
\maketitle
\begin{abstract}
Past work on personality detection has shown that frequency of lexical categories such as first person pronouns, past tense verbs, and 
 sentiment words have significant correlations with personality traits. 
In this paper, for the first time, we show that fine affect (emotion) categories
such as that of excitement, guilt, yearning, and admiration are significant indicators of personality.
Additionally, we perform experiments to show that the gains provided by the fine affect categories are not obtained by using
coarse affect categories alone or with specificity features alone.
We employ these features in five SVM classifiers
for detecting five personality traits through essays.
We find that the use of fine emotion features leads to statistically significant improvement over a competitive baseline,
whereas the use of coarse affect and specificity features does not.

\end{abstract}

\section{Introduction}
Personality has significant impact on our lives---for example, on job performance \cite{Tett91} and
inter-personal relations \cite{White04}. 
The five-factor or the big five model of personality
describes personality along the dimensions of extroversion vs.\@ introversion (sociable, assertive vs.\@ aloof, shy),
 neuroticism vs.\@ emotional stability (insecure, anxious vs.\@ calm, unemotional),
agreeability vs.\@ disagreeability (friendly, cooperative vs.\@ antagonistic, fault-finding),
conscientiousness vs.\@ unconscientiousness (self-disciplined, organized vs.\@ inefficient, careless),
openness to experience vs.\@ conventionality (intellectual, insightful vs.\@ shallow, unimaginative) \cite{Mairesse07}.

Traditionally, researchers determine personality through specific questionnaires.
However, automatically identifying personality from free-form text is far more desirable. 
Past work has shown that certain features such as the use of first person pronouns ({\it I, we}), use of
words relevant to social processes ({\it chat, friend}), use of past tense ({\it had, was}), and the use of
certain emotion words ({\it hate, angry}) have significant correlations
with different personalities \cite{PennebakerK99,Mairesse07}.
Many of these approaches relied on small manually created lexicons of
sentiment and other lexical categories such as lists of pronouns,
determiners, articles, social words, past tense verbs, and so on.
Interestingly, word ngrams, one of the most widely used features in natural language processing and especially useful for text categorization by topic,
tend not to be very helpful in personality detection.

In this paper, for the first time, we show that lexical categories corresponding to fine-grained
emotions such as excitement, guilt, yearning, and admiration are significant indicators of personality.
Personality has a known association with emotion. Emotions are considered to be more transient phenomenon
whereas personality is more constant. 
Plutchik \shortcite{Plutchik62} argues
that the persistent situations involving such emotions produce persistent traits or personality.
Past work has used small lexical categories pertaining to a few basic emotions such as anger, joy, and sadness.
We believe that personality detection can benefit from a much larger lexical database with information about many different fine-grained emotions.

Further, we wanted to determine whether the gains obtained by fine affect categories are truly
because of affectual grouping of words into fairly specific categories. Thus we set up
comparative experiments using coarse affect features and word specificity features.
We explore three affect and specificity features that draw from large automatically created lexicons:
(1) {\it the NRC Hashtag Emotion Lexicon (Hashtag Lexicon, for short)}: a lexicon of word associations with 585 emotions,
(2) {\it the Osgood dimensions lexicon}: a lexicon of word evaluativeness, potency and activity, and
(3) {\it the specificity lexicon}: a lexicon of word specificity captured in the form of information content.
We created the Hashtag Lexicon from about 775,000 tweets with emotion-word hashtags, following the idea of Mohammad \shortcite{Mohammad12}.
In contrast with the Hashtag Lexicon, which has fine-grained affect categories, the Osgood Lexicon has coarse affect categories.
It was created by Turney \shortcite{TurneyL03} for sentiment analysis. We explore its use for personality detection.
The specificity of a word is a measure of how general or specific the concept being referred to is.
We create a word-level specificity lexicon using Pedersen's precomputed scores of WordNet synset specificities.\footnote{http://wn-similarity.sourceforge.net.}

We employ the affect and specificity features in  state-of-the-art SVM classifiers
and detect personalities of people through their essays.
The Essays dataset we use was collected by Pennebaker and King \shortcite{PennebakerK99} 
and consists of 2469 essays (1.9 million words) by psychology students.
The dataset was provided as part of a shared task in the Workshop on Computational Personality Detection.\footnote{http://mypersonality.org/wiki/doku.php?id=wcpr13}
Personality was assessed by asking the students to respond to a Big Five Inventory Questionnaire \cite{John99}.
We find that the use of fine emotion features leads to statistically significant improvement over a competitive baseline,
whereas the use of coarse affect and specificity features does not.

\section{Related Work}

Pennebaker and King \shortcite{PennebakerK99}
used lexical categories from Linguistic Inquiry and Word Count (LIWC) 
to identify linguistic correlates of personality.\footnote{http://www.liwc.net} They showed, for example,
that agreeability is characterized with more positive emotion words and fewer articles and
that neurotism is characterized with more negative emotion words and more first-person
pronouns.
Mairesse et al.\@ \shortcite{Mairesse07} improved on these features
and distribute their system online.\footnote{http://people.csail.mit.edu/francois/research/personality/\\recognizer.html}
We use all of their features to create our baseline classifier---{\it the Mairesse baseline}.
Some of these features are listed below:
word count, words per sentence, type/token ratio, words
longer than six letters, negations, assents, articles,
prepositions, numbers, pronouns (first person, second person, third person),
emotion words, cognition words ({\it insight, tentative}), sensory and perceptual words ({\it see, hear}),
social processes words ({\it chat, friend}), time words, space words, motion words,
punctuations, and swear words.
Both Pennebaker and King \shortcite{PennebakerK99} and Mairesse et al.\@ \shortcite{Mairesse07} worked with
the Essays dataset. More recently, there is also work on personality detection from blogs \cite{Yarkoni10},
Facebook posts \cite{Kosinski13},
and Twitter posts and follower network \cite{Qiu12}.
There also exist websites that analyze blogs and display the personality types of the authors.\footnote{http://www.typealyzer.com}

%

\section{Proposed Features for Personality Detection}

\subsection{Fine Affect Categories}


The NRC Hashtag Emotion Lexicon \cite{Mohammad12,MohammadKa13} has word--emotion association scores for 585 emotions.
A list of 585 emotion-related hashtags (e.g., \#love, \#annoyed,
\#pity) was compiled from different sources. 
Then, about 775,000 tweets
containing at least one of these hashtags were collected from Twitter.
Simple word counts were used
to calculate pointwise mutual information (PMI) between an emotional hashtag and a word
appearing in the tweets. 

The PMI represents a degree of association between the word and
emotion, with larger scores representing stronger associations. 
The lexicon (version 0.1)
contains around 10,000 words with associations to 585 emotion-word hashtags. 
We used the NRC Hashtag Lexicon by creating a separate feature for each
emotion-related hashtag,  resulting in 585 emotion features. The values of these
features were taken to be the average PMI scores between the words in an
essay and the corresponding emotion-related hashtag.

In order to compare with coarse-grained emotion features, we used the NRC Emotion Lexicon \cite{MohammadT10}.
 The lexicon is comprised of 14,182 words manually annotated with eight basic
 emotions (anger, anticipation, disgust, fear, joy, sadness, surprise, trust).
  Each word can have zero, one, or more
 associated emotions.
We created eight features from this lexicon in the same manner
as the Hashtag Lexicon features.

\subsection{Coarse Affect Categories}
Osgood et al.\@ \shortcite{Osgood57} asked human subjects to rate words on various scales such as complete--incomplete,
harmonious--dissonant, and high--low. They then performed a factor analysis of these ratings to discover
that most of the variation was due to three dimensions: evaluativeness ({\it good--bad}), activity ({\it active--passive, large--small}),
and potency ({\it sharp--dull, fast--slow}).
Turney and Littman \shortcite{TurneyL03} proposed a method to automatically calculate a word's evaluativeness score using a vector
space model and word--word co-occurrence counts in text.
Turney later generated lexicons of word--evaluativeness scores and additionally lexicons of word--activity and word--potency
scores for 114,271 words from WordNet.
We used these lexicons and computed the average evaluativeness, activity, and potency scores of the words in an essay. 

\subsection{Specificity}
Gill and Oberlander \shortcite{GillO02}, and later Mairesse et al.\@ \shortcite{Mairesse07}, show that 
people with a neurotic personality tend to use concrete words more frequently.
Inspired by this, we explore if people of a certain personality type tend to use
terms with high specificity.
The specificity of a term is a measure of how general or specific the referred concept is.
For example, {\it entity} is a very general concept whereas {\it ball-point pen} is a very specific concept.

Resnik \shortcite{Resnik95} showed that specificity or information content of WordNet synsets can be accurately determined by using
corpus counts. 
Pedersen pre-computed information content scores for 82,115 WordNet noun synsets and 13,708 verb synsets using the British National Corpus (BNC). 
We created a word-level information content lexicon by first mapping the words to their synsets, and then assigning the words with information content scores
of the corresponding synsets. If a word is associated with more than one synset, then the synset with the highest information content
is chosen. The final lexicon had 66,464 noun entries and 6,439 verb entries.
We computed the average information content of the words in an essay and used it as a feature in our machine learning system.

\section{Automatically Identifying Personality}

We trained five Support Vector Machine (SVM) classifiers for each of the five
personality dimensions. SVM is a state-of-the-art learning
algorithm proven to be effective on text categorization tasks and robust on
large feature spaces. 
In each experiment, the results were averaged over three-fold stratified cross-validation. We used the LibSVM
package \cite{CC01a} with a linear kernel. 
Each essay was represented by the following groups of features:
\begin{enumerate}[a.] 
	\item Mairesse Baseline (MB): This is the complete set of features used by Mairesse \shortcite{Mairesse07}. (Described earlier in the Related Work Section.)
	\item Token unigrams: Frequencies of tokens divided by the total number of tokens in an essay.
	\item Average Information Content (AIC): Average information content of the essay, calculated using the Specificity Lexicon.
	\item Features from coarse affect categories (CoarseAff): Average potency of the essay, average evaluativeness of the essay, and the average activity score of the essay calculated using
	the Turney lexicons.
	\item Features from basic emotion categories (BasicEmo): Average of the emotion association score for each of the 8 emotions in the NRC Emotion Lexicon.
	\item Features from fine emotion categories (FineEmo): Average of the emotion association score for each of the 585 emotions in the NRC Hashtag Lexicon.	
\end{enumerate}

Upon classification, the results were compared with the gold labels of yes or no for each of the five
personality dimension to determine precision, recall, and F1-score.
Table~\ref{tab:results} shows the macro-average F1-scores of the yes and no labels
for the five personality classes extroversion (EXT), neurotism (NEU),
agreeability (AGR), conscientiousness (CON), and openness (OPN). We also present the results for a simple baseline classifier that always predicts the majority class. 

Observe that the biggest gains over the Mairesse baseline are provided by the 585 fine-grained
emotion categories of the Hashtag Lexicon (row f). Further, they lead to improvements in the detection of all five personality classes. 
To confirm the significance of these results, we repeated the experiments 10 times and compared the scores with a paired t-test. 
We found that the improvements the Hashtag Lexicon features offers over the Mairesse baseline are statistically significant with 99\% confidence for three out of five classifiers: EXT, CON, and OPN. 
Note that using only eight basic emotion categories of the NRC emotion lexicon leads to much smaller improvements over MB (row e).
This is despite the fact that the NRC Lexicon has more entries than the Hashtag Lexicon.
Note also that adding unigram features over the Mairesse baseline does not improve the results (row b has similar values as in row a).
This suggests that the Hashtag Lexicon is providing improvements not because of its vocabulary,
but rather because of the way it groups the vocabulary into nuanced emotion categories.

Adding average information content to the Mairesse baseline improves results for the
EXT class, but the improvement is not statistically significant. 
Using information content features of nouns alone or verbs alone led
to similar results.
The coarse affect features (d.\@ rows) provide a slight improvement for the EXT, CON, and OPN classes,
but again the improvements are not significant.


Row h of Table 1 shows the results obtained when using hashtag lexicon features alone (no Mairesse Baseline features).
Observe that these numbers are comparable and sometimes (for CON and OPN) even better than the MB features. 

\begin{table}[t]
\begin{center}
\resizebox{0.49\textwidth}{!}{
\begin{tabular}{llllll}
\hline
                                                                                                                &EXT   &NEU   &AGR   &CON   &OPN\\ \hline
                               Majority Classifier                         &             51.70        &             49.90        &             53.10        &             50.80        & 51.50\\[3pt]
							SVM Classifier & & & & &\\
                                a. MB  										 &54.78  		&58.09  		&54.19  		&55.05  &59.56\\ 
                                b. MB + Unigrams                             & 55.57   & 57.94    & 53.63   &55.23  &59.22\\
                    c. MB + AIC                                              &54.86  	&58.05   	&54.14  		&54.96  &59.47\\[3pt]
                d. MB + CoarseAff                                             &                             &                             &                             &                             &\\
                    $\,\,\,\,\,\,\,\,\,\,$ Activity                          &55.45  		&57.62  		& 53.48  	&54.94  &59.23\\
                    $\,\,\,\,\,\,\,\,\,\,$ Evaluative                        &55.25  		&57.66  		&53.02  		& 55.84  & 59.71\\
                    $\,\,\,\,\,\,\,\,\,\,$ Potency                           &54.64   		&57.37  		&53.25  		& 55.38  & 59.56\\
                    $\,\,\,\,\,\,\,\,\,\,$ All three                         &55.32   		& 58.10   	&52.95  		& 55.48  & 59.88\\[3pt]
                    e. MB + BasicEmo                     					 & 55.50 & 58.23 	&53.04 			& 55.53 & 59.10\\
                    f. MB + FineEmo                                          &{\bf 56.28}  &58.25   	& 54.20  	& {\bf 56.56}  &{\bf 60.61}\\
                    g. MB + c + d + f                                        &{\bf 56.28}  &58.15   	&53.90   	&{\bf 56.35}  &{\bf 60.57}\\
                    h. FineEmo alone																								& 54.68 & 55.74 & 54.02 & {\bf 56.46} & {\bf 60.43}\\
\hline
\end{tabular}
}
\end{center}
\caption{\label{tab:results} Macro-averaged F1-score of automatic essay classification into the big five dimensions of personality.
All improvements over MB that were statistically significant are shown in bold.}
\end{table}

\section{Discussion}
The fact that unigram features are not as helpful as in some other tasks such as classification of text by topic,
is one of the reasons personality detection is a relatively hard problem.
Nonetheless, the fine-grained emotion features from the Hashtag Lexicon provided statistically significant gain
over the baseline. In contrast, coarse affect features and specificity features failed to
provide significant improvements. This suggests that fine affect categories contain useful
discriminating information not present in coarse affect categories or simple specificity features.

In order to identify which of the 585 emotions had the most discriminative
information, we calculated information gain of each of 585 emotion features. (Decision tree learners
use information gain to determine the sequence of nodes in the tree.)
Table \ref{tab:gain} shows the top ten emotion categories with the highest gain for the five personality dimensions.
Observe that most of the emotions seem to be reasonable indicators of the corresponding personality trait.
Note that the columns include emotions that are indicative of either of the two ends of the personality dimensions 
(for example, the emotions in column EXT are associated with either extroversion or introversion).
Observe also that some of these emotions are very close to the basic emotions of happiness and sadness, 
but many are emotions felt at relatively specific situations, such as guilt, excitement, anxiety, and shame.

The five terms most associated with the lexical categories of {\it \#possessive} and {\it \#apart}
(the two most discriminative emotion categories for EXT)
are shown below:\\
\indent {\bf \#possessive:} 1. {\it possessive}: 7.228    2. {\it hottie}: 6.448    \\\indent  3. {\it tense}: 5.911 4. {\it lover}: 5.213 5. {\it mine}: 4.141.\\
\indent {\bf \#apart:} 1. {\it apart}: 4.6 2. {\it tear}: 4.065  3. {\it miss}: 2.341  \\\indent 4. {\it fall}: 2.085  5. {\it heart}: 1.63.\\
\noindent The numbers next to the words are their PMI scores with the emotion word hashtag.
Observe that the terms in the \#possessive category tend to be used more often by an extrovert,
whereas the terms in the \#apart category tend to be associated more with introverts.



\begin{table}[t]
\begin{center}
 \resizebox{0.47\textwidth}{!}{
\begin{tabular}{lllll}
\hline
        EXT   		&NEU   			&AGR   		&CON   			&OPN\\ \hline
		possessive	&guilt			&happy		&excited		&anxious\\
		apart      &eager       	&anger      &apprehensive   &delighted\\
		happy   &interested     &homesick   &anger       	&blah\\ 
		cherish    &keen       	&giddy       &hate       	&exhausted\\ 
      admiring   &helpless       &chaotic     &ashamed       &sweet\\
      impaired   &passion       	&heartbroken &giddy       	&tired\\
      jealousy   &unhappy       	&sweet       &partial       &lonely\\
      gleeful    &insignificant  &neglected   &disturbed     &nervous\\
      vibrant    &timid       	&loving      &wrecked       &ecstatic\\
     huggy      &anticipation   &lonely      &needed       	&wrecked\\ 

\hline
\end{tabular}
}
\end{center}
\caption{\label{tab:gain} Top ten hashtag emotion categories with highest information gain for personality classification.}
\end{table}

\section{Conclusions}
It is well-established that there is relation between emotions and personality, however automatic 
personality detection has thus far used other features such as lexical categories of pronouns and articles.
 In this paper, for the first time, we showed that lexical categories corresponding to fine-grained
 emotions such as excitement, guilt, yearning, and admiration are significant indicators of personality.
 We performed experiments using three 
 large automatically created lexicons
of fine emotion categories, coarse affect categories, and word information content.
We developed state-of-the-art SVM classifiers using a set of previously successful features, and added to it the three new sets of features. 
 All three sets of features improved performance of one or more classifiers over a strong baseline of previously successful features.
 The improvements obtained with the fine emotion categories (the NRC Hashtag Emotion Lexicon) were particularly significant.
We believe that 
even further gains may be obtained by combining
sophisticated sentence-level emotion analysis with personality detection.
The improvements obtained using coarse affect categories and information content were not statistically significant,
but it is still possible that personality detection can benefit from a more creative use of these features.
All resources created by the authors and used in this research effort, including the Hashtag Lexicon, are freely available.\footnote{Email Saif Mohammad (saif.mohammad@nrc-cnrc.gc.ca).}


\bibliography{references}

\end{document}